\pdfoutput=1

\documentclass[11pt]{article}
\usepackage[preprint]{acl}
\usepackage{pgfplots}
\usepackage{pgfplotstable}
\pgfplotsset{compat=1.17}
\usepackage{tikz}
\usepackage{booktabs}
\usetikzlibrary{automata,positioning}
\usepackage{xcolor}
\usepackage{listings}

\lstdefinelanguage{pseudo}{
    morekeywords={def, return, if, else},
    morekeywords={[2]select, aggregate, indicator},
    sensitive=true,
    morecomment=[l]{//},
    morestring=[b]",
}

\usepackage{times} 
\usepackage{latexsym}
\usepackage{amsmath}
\usepackage{amsfonts}
\usepackage{algorithm}
\usepackage{algpseudocode}
\usepackage[T1]{fontenc}

\usepackage[utf8]{inputenc}

\usepackage{microtype}

\usepackage{inconsolata}
\usepackage{tcolorbox}
\usepackage{graphicx}
\usepackage{subcaption}
\usepackage{float}
\title{Emergent Stack Representations in Modeling Counter Languages Using Transformers}

\author{
  \textbf{Utkarsh Tiwari\textsuperscript{1, *}}, 
  \textbf{Aviral Gupta\textsuperscript{1, *}}, 
  \textbf{Michael Hahn\textsuperscript{2, \textdagger}} \\
  \textsuperscript{1}Birla Institute of Technology and Science, Pilani \\
  \textsuperscript{2}Saarland University \\
  \small
  \textbf{Correspondence:} \href{mailto:f20220052@pilani.bits-pilani.ac.in}{\{f20220052, f20220097\}@pilani.bits-pilani.ac.in} \\
}

\begin{document}
\maketitle
\begin{abstract}
Transformer architectures are the backbone of most modern language models, but understanding the inner workings of these models still largely remains an open problem. One way that research in the past has tackled this problem is by isolating the learning capabilities of these architectures by training them over well-understood classes of formal languages. We extend this literature by analyzing models trained over counter languages, which can be modeled using counter variables. We train transformer models on 4 counter languages, and equivalently formulate these languages using stacks, whose depths can be understood as the counter values. We then probe their internal representations for stack depths at each input token to show that these models when trained as next token predictors learn stack-like representations. This brings us closer to understanding the algorithmic details of how transformers learn languages and helps in circuit discovery.
\end{abstract}

\section{Introduction}

\def\thefootnote{*}\footnotetext{Equal contribution}\def\thefootnote{\arabic{footnote}}
\def\thefootnote{\textdagger}\footnotetext{Senior author}\def\thefootnote{\arabic{footnote}}

Modern day language models (LMs) are increasingly capable of capturing complex sequential and linguistic patterns, achieving strong performance across diverse natural language as well as synthetic tasks. Despite their impressive capabilities and substantial amounts of effort and progress, the inner workings of these models still remains opaque and un-interpretable to a substantial extent. For instance, we know that transformer models of sufficient complexity can learn, for example, to perform modular arithmetic, but recovering the exact algorithm used by these models to perform this task remains a challenge \citep{nandaprogress}. Building on this line of inquiry, our research focuses on formal languages, specifically examining models trained on counter languages—a class of languages formally modeled using stack memory. We demonstrate that these models develop internal representations that effectively mimic stack structures, providing insights into how they process and generate such languages.\footnote{Importantly, we do not make any claims about the causality of these stacks. This is further discussed in the Section 5.} Interpreting, understanding and reasoning about the algorithms and circuits within language models (LMs) remains a largely open problem, with use cases in AI safety, alignment, and performance.

\begin{figure}[t]
    \centering
    \includegraphics[width=\linewidth]{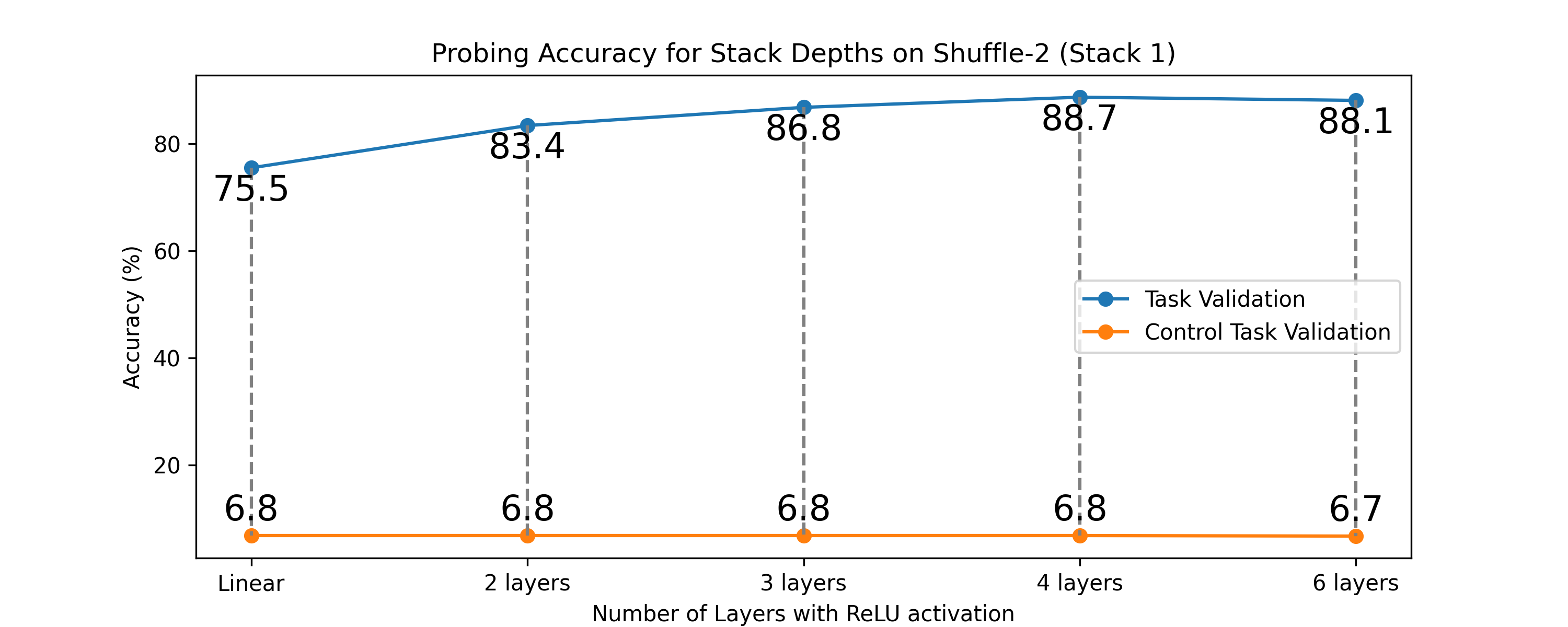}
    \caption{High probing accuracy for the counter values at each token on a model trained to recognize counter languages as a next token predictor shows that the models learn a stack-like structures in their internal representation which is helpful in reasoning about and interpreting the learned algorithm to achieve this task.}
    \label{fig:example}
\end{figure}

 \textbf{Formal Languages:} Formal languages provide a useful testbed to isolate and investigate the learning properties of transformers and their failure cases, since these languages have precise mathematical properties that the model can be tested on \citep{ackerman2020surveyneuralnetworksformal}. This literature contains both empirical results \citep{strobl-etal-2024-formal, Bhattamishra2020OnTA} as well as theoretical analysis of cognitive biases and learning bounds over these formal tasks \citep{hahn-rofin-2024-sensitive, pérez2019turingcompletenessmodernneural, zhou2023algorithmstransformerslearnstudy, hahn-2020-theoretical}.

This line of inquiry is particularly significant because all algorithmic tasks can be reduced to a language within a specific class of formal languages. For instance, arithmetic in Polish notation can be modeled as a single-counter, non-regular context-free language, solvable with a stack memory but not without it. Thus, robust theoretical and empirical insights into formal languages contribute to advancing the capabilities of language models \citep{zhang2024transformerbasedmodelsperfectlearning}. Furthermore, natural languages exhibit features that can be approximately mapped to certain classes of formal languages, often displaying recursive structures akin to those found in formal systems \citep{kornai-1985-natural, Jger2012FormalLT}.

\textbf{Mechanistic Interpretability and Probing Classifiers:} The field of Mechanistic Interpretability (MI)    aims to extract the reasoning and interpretable structures present inside these models. Among other things, MI deals with identifying \textit{features} inside language models \citep{rai2024practicalreviewmechanisticinterpretability}. A feature is a human-interpretable property of the model’s activations on specific inputs. This leads to an understanding of features as meaningful vectors in the activation space of a model. A widely used approach to understanding model structures through the aforementioned features involves \textit{probing} these models by linking internal representations or activations with external properties of the inputs. This is done by training a classifier on these representations to predict specific properties. Known as probing classifiers \citep{belinkov-2022-probing}, this framework has become a key analysis tool in numerous studies of NLP models and the methodology that we use to understand the internal structure of the trained models on these formal languages. This direction is also motivated by other works which have had success in retrieving coherent internal structures like world models in natural language \citep{li-etal-2021-implicit, abdou-etal-2021-language} and toy/synthetic setups \citep{elhage2022toymodelssuperposition, li2024emergentworldrepresentationsexploring, vafa2024evaluatingworldmodelimplicit}. To our knowledge, this work is the first to leverage probing classifiers to analyze models trained on formal languages. 

To more formally define the training objective of the probes, let \( f: x \mapsto \hat{y} \) denote a language model, trained on a formal language, with performance measured by the language modeling task of auto-regressive next-token prediction. The model \( f \) generates intermediate representations \( f_l(x) \) at layer \( l \), called embeddings. A probing classifier \( g: f_l(x) \mapsto \hat{z} \) maps these embeddings to a property \( z \) (e.g., stack depth, part-of-speech), trained and evaluated on dataset \( D_P = \{(f_l(x), z^{(i)})\} \), which is formed by pairing each token's embedding with its property in that sequence. The performance of the probing classifier depends on two key factors: the probe's ability to map embeddings to the target property and the original model \( f \)'s ability to generate information-rich embeddings by effectively learning the next-token prediction task. If the classifier achieves high performance, it indicates that the model has learned information relevant to the property being probed. This setup serves as a proxy for examining the internal structure of the transformer, by probing for implicit properties of the dataset it was trained on. However, the choice of property must be carefully considered to ensure meaningful and interpretable results. In our case, we adopt a more formal and structured approach by leveraging counter languages, which provide a well-defined and rigorous framework for probing. Formal languages offer a clear and systematic property to probe, grounded in their precise modeling and theoretical foundations.

\section{Related Work}

\textbf{Transformers and Formal Language Learnability:}
Transformers dominate sequence modeling tasks, but their ability to model FL requiring structured memory, like counters or stacks, is only incompletely understood. While theoretically Turing-complete \citep{pérez2019turingcompletenessmodernneural} and universal approximators of sequence functions \citep{yun2020transformersuniversalapproximatorssequencetosequence}, practical learnability of FL is less understood. For example, \citet{hahn-2020-theoretical} showed Transformers struggle with Parity and Dyck-2 in the asymptotics of unbounded sequence length. Empirical studies, such as \citet{Bhattamishra2020OnTA} and \cite{Strobl_2024}, demonstrate Transformers can learn Dyck-1 and Shuffle-Dyck, suggesting they can simulate counter-like behavior. These findings highlight the need for further investigation into Transformers' ability to model FL.

\textbf{Probing Internal Representations:}
Probing classifiers have become a key tool for understanding neural model representations. By training simple classifiers on intermediate activations, researchers infer whether specific properties are encoded. For example, \citet{voita2019analyzingattention} studied attention heads, while \citet{rogers2020primer} and \citet{coenen2019visualizingmeasuringgeometrybert} explored intermediate layer information. In FL, probing has been used to analyze models trained on tasks like arithmetic and syntactic parsing \citep{li-etal-2021-implicit, abdou-etal-2021-language}. Notably, \citet{elhage2022toymodelssuperposition} and \citet{li2024emergentworldrepresentationsexploring} showed probing can reveal emergent structures in synthetic setups. Our work extends this by probing models trained on counter languages, providing insights into stack-like representations learned by Transformers.

\textbf{Enforcing Stack Structures in Transformers:}
Several works have explored explicitly incorporating stack-like structures into neural models to handle hierarchical patterns. For example, \citet{stackrnn} proposed a neural stack for RNNs, enabling pushdown automata simulation. Similarly, \citet{suzgun2019memoryaugmentedrecurrentneuralnetworks} introduced a differentiable stack for learning Dyck languages. In Transformers, papers like \citet{transformerstack} and \citet{stackattention} explored external memory modules to enhance long-range dependency modeling. While promising, these approaches require significant architectural changes. Our work focuses on whether Transformers can implicitly learn stack-like representations without explicit constraints, offering a more interpretable approach to modeling FL.


In summary, our work builds on these foundations by analyzing Transformers' ability to model counter languages and probing their internal representations for stack-like structures. By bridging FL theory and mechanistic interpretability, we aim to advance understanding of how Transformers learn and generalize algorithmic patterns.

\section{Problem Setup And Architecture}
\subsection{Counter Language Modeling}

Counter Languages are languages modeled using counter machines, which are DFAs with counter variables which can be incremented, decremented, and set to 0. We focus our work on the Dyck language, which is a well studied class of counter languages, and the $k$-Shuffles of Dyck-$1$. 

\textbf{Dyck-1} is the set of well-formed parenthesis strings, is a context-free language which requires 1 counter to model it. Over the alphabet $\Sigma = \{(, )\}$ the production rules are:
\[
S \to 
\begin{cases} 
    \epsilon \\
    (S) \\
    SS
\end{cases}
\]

\textbf{Shuffle} is the binary operation $||$ on two strings which interleaves the two string in all possible ways. Inductively:
\begin{itemize}
    \item \( u \odot \epsilon = \epsilon \odot u = \{u\} \)
    \item \( \alpha u \odot \beta v = \alpha (u \odot \beta v) \cup \beta (\alpha u \odot v) \)
\end{itemize}
for any $\alpha, \beta \in \Sigma$ and $u, v \in \Sigma^*$\footnote{The $*$ is the Kleene Star operation}. For example, the shuffle of $ab, cd$ = \{$abcd, acbd, acdb, cabd, cadb, cdab$\}. The Shuffle operation can be extended to apply over languages $\mathcal{L}_1$ and $\mathcal{L}_2$ as:
\[
\mathcal{L}_1 \odot \mathcal{L}_2 = 
\bigcup_{\substack{u \in \mathcal{L}_1, \\ v \in \mathcal{L}_2}} u \odot v
\]

We use Shuffle-$k$ to denote the Shuffle of $k$ Dyck-1 languages, each with vocabulary $\Sigma_i$ such that $\bigcap_{i\in [1, k]} \Sigma_i = \emptyset$, i.e. disjoint vocabularies\footnote{Hence, Shuffle-$1$ is the same as Dyck-$1$}. 

Similar to \citet{Bhattamishra2020OnTA} Shuffle-2 is the shuffle of Dyck-1 over alphabet $\Sigma$ = \{(, )\} and another Dyck-1 over the alphabet $\Sigma$ = \{[, ]\}. Hence the resulting Shuffle-2 language is defined over alphabet $\Sigma$ = \{[, ], (, )\} and contains words such as ([)] and [((])) but not ])[(.

In our experiment setup we consider 4 counter languages: Dyck-1, and Shuffle-2, Shuffle-4, and Shuffle-6. For all these languages we follow the training details from \citet{Bhattamishra2020OnTA}

\subsection{Language Modeling Architecture and Training Setup}

\textbf{Model Architecture.} We employ an encoder-only transformer model with a linear decoder layer (language-modeling head) for sequence processing tasks, following the setup from \citet{Bhattamishra2020OnTA}. Let \( \mathcal{V} \) denote the alphabet of the language, where each symbol \( v \in \mathcal{V} \) is treated as a unique token. Tokenization is a mapping \( \tau: \mathcal{V} \to \mathbb{Z} \), assigning each token to a unique integer ID. The encoder maps input tokens \( x = (x_1, \dots, x_T) \) to dense embeddings \( E = (e_1, \dots, e_T) \), where \( e_i \in \mathbb{R}^{d_{\text{model}}} \), augmented with positional encodings \( P = (p_1, \dots, p_T) \) to encode sequence order. The combined embeddings \( X = E + P \) are processed by a stack of \( n_{\text{layers}} \) transformer encoder layers. 
Each layer consists of:
{\normalsize
\begin{enumerate}
    \item A multi-head self-attention mechanism with a causal mask \( M \), ensuring
    \[
        \text{Attention}(Q, K, V) = \text{softmax}\left(\frac{QK^T}{\sqrt{d_k}} + M\right)V
    \]
    where \( M_{ij} = -\infty \) if \( i < j \) and \( 0 \) otherwise.
    \item A feed-forward network (FFN) with \( d_{\text{ffn}} \) hidden units, applied position-wise.
\end{enumerate}
}
The output of the final encoder layer \( H = (h_1, \dots, h_T) \) is projected into the output space through a linear decoder layer \( W_{\text{out}} \in \mathbb{R}^{|\mathcal{V}| \times d_{\text{model}}} \), followed by a sigmoid activation \( \sigma \), yielding final output \( \hat{y} = \sigma(W_{\text{out}} H) \). Residual connections and layer normalization are applied throughout to stabilize training. We restrict the model size to prevent overfitting and facilitate interpretability.

This architecture is designed to process sequences auto-regressively, with the self-attention mechanism ensuring that each token \( x_i \) attends only to itself and preceding tokens \( x_{j \leq i} \). The model used in our experiments has a hidden dimension of 64 and the embedding dimension of 32. With only 1 layer deep transformer and 4 attention heads.
\\
\textbf{Training Setup.} The model is trained on the next-token prediction task, where it processes an input sequence \( s = (s_1, \dots, s_n) \) and predicts the set of valid characters for the next step \( s_{i+1} \) given the subsequence \( s_{1:i} \). The model outputs a distribution over \( \mathcal{V} \), represented as a \( k \)-dimensional vector \( \hat{y} \in [0, 1]^{|\mathcal{V}|} \), where \( k = |\mathcal{V}| \). The ground truth is a \( k \)-hot vector \( y \in \{0, 1\}^{|\mathcal{V}|} \), indicating valid next characters. The training objective is to minimize the mean squared error (MSE) between \( \hat{y} \) and \( y \).

During inference, predictions are obtained by thresholding \( \hat{y} \) at 0.5. A sequence is considered correctly recognized if all predictions match the ground truth at every step. The model is trained using RMSProp optimizer with a learning rate of \( 5 \times 10^{-3} \) for 25 epochs with a batch size of 32 and absolute positional encodings.
\begin{figure}
    \centering
    \includegraphics[width=1\linewidth]{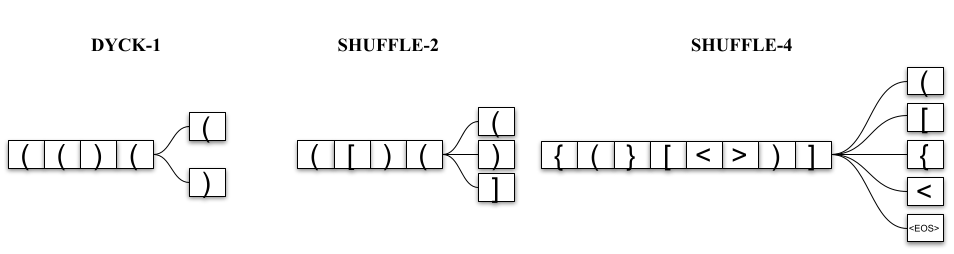}
    \caption{The model is trained to predict the set of all valid next tokens.}
    \label{fig:enter-label}
\end{figure}

\subsection{Probing Setup}
\label{sec:probing}

Shuffle-$k$ can be modeled using $k$ counter variables. Let $s$ be the input string of length $n$, and let $\text{paren}_i$ denote the $i$-th pair of parentheses with opening symbol $\text{open}_i$ and closing symbol $\text{close}_i$ for $1 \leq i \leq k$. During iterating over $s$ and increasing the value of counter$_i$ when $\text{open}_i$ is encountered and decreasing it when $\text{close}_i$ is encountered, if any counter value becomes negative then we can say $s \notin \text{Shuffle-}k$. At the end of the string if the values of all counters are 0 then $s \in \text{Shuffle-}k$.

Algorithmically, Shuffle-$k$ can be modeled using $k$ stacks, and pushing $\text{open}_i$ in stack$_i$ and popping when $\text{close}_i$ is encountered. If any stack$_i$ is empty when close$_i$ is encountered then $s \notin \text{Shuffle-}k$ and if at the end all stacks are empty then $s \in \text{Shuffle-}k$. In this formulation the depth of stack$_i$ corresponds to counter$_i$. Hence, for a Shuffle-$k$ language, we train $k$ different probing models, one for each stack.

We probe the trained models for the depth of stacks at each input token to determine if the models learns the counter representation of these languages. We train simple feed forward networks of varying depths and complexity as multi class classifiers with ReLU activations on the internal representations from the trained model to predict the depth of stack being probed for.

The probing dataset is constructed by sampling sequences from the training corpus of the language model. It consists of 10,000 samples, each with lengths ranging between 2 and 50 tokens. The dataset is split into 8,000 samples for training and 2,000 for validation. Each sample is represented as a pair comprising the transformer encoder's embedding and the corresponding probed value. The embedding is extracted from the output of the last encoder layer of the language model and has a dimension of \(d_{\text{model}}\). Specifically, for an input sequence of length \(T\), the encoder produces an output of dimension \(T \times d_{\text{model}}\). We slice this output along the sequence length dimension to obtain a single embedding of size \(1 \times d_{\text{model}}\) for each token, which is then included in the probing dataset. This process is repeated for all samples in the dataset, ensuring that each entry captures the relevant encoder representation for probing tasks.

Keeping in the with the best practices of the literature \citep{hewitt2019designinginterpretingprobescontrol} we also create a control task by randomizing the target values for the same input set. The difference between the accuracy on the main task and the control task is called the \textit{selectivity}, and high values signify that the probing model is not learning spurious correlations, or memorizing the training data.

To further verify our probing methodology, we compile a model $M$ using \textbf{Tracr} \citep{NEURIPS2023_771155ab} from the \textbf{RASP} \citep{weiss2021thinking} program for Dyck-1 and use the same probing setup $P$ on this model. Since, $M$ is not trained using data, but is instead compiled using a known algorithm it can be used to test if $P$ is effective at detecting stack-like features. The results of $P$ on $M$ are positive and are included in the Appendix.

\section{Results}
\begin{figure}[t] 
    \centering
    \includegraphics[width=\linewidth]{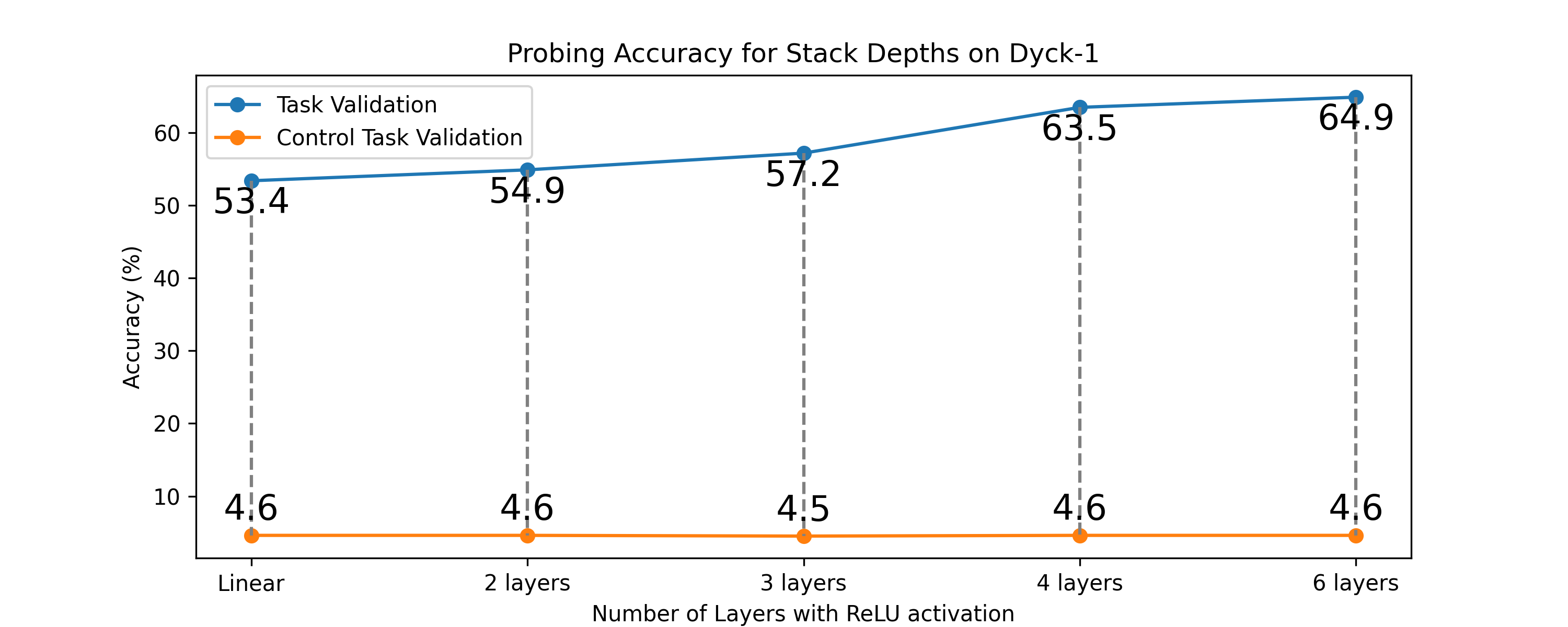} 
    \vskip\baselineskip 
    \includegraphics[width=\linewidth]{Shuffle2S1.png}
    \vskip\baselineskip
    \includegraphics[width=\linewidth]{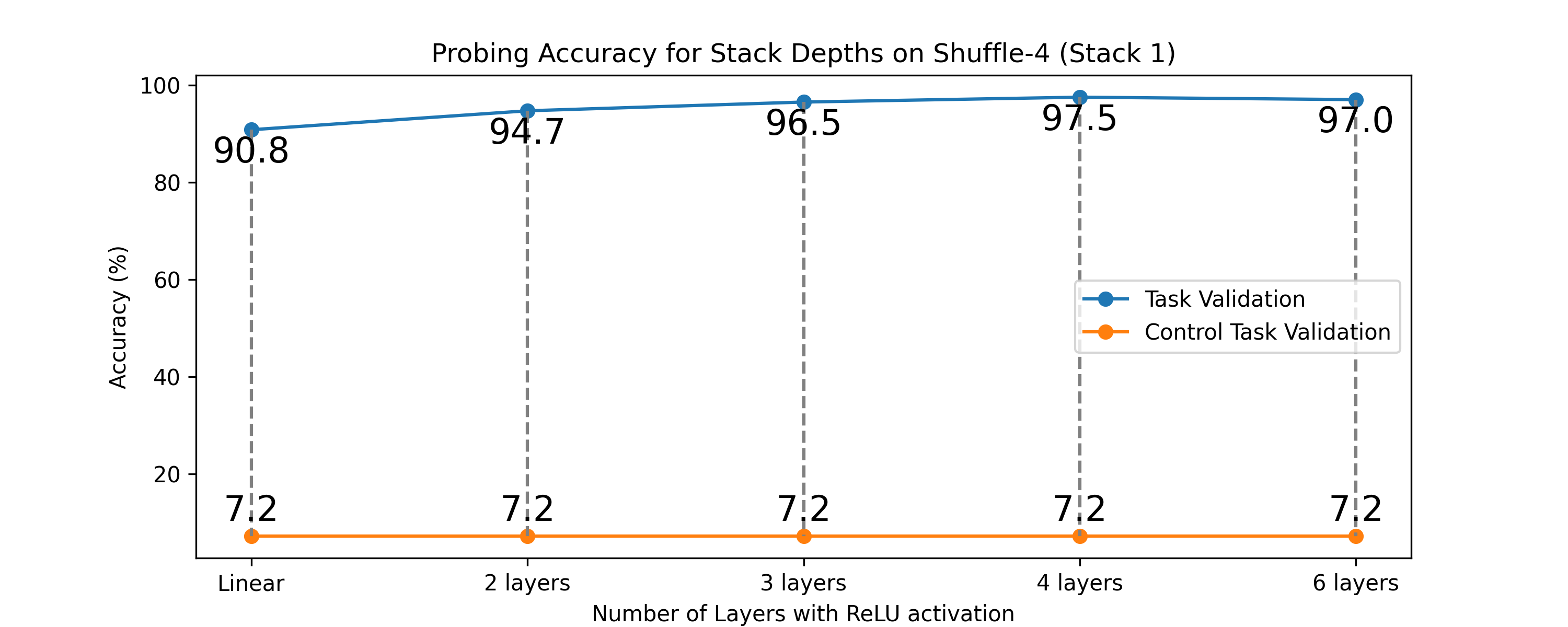}
    \vskip\baselineskip
    \includegraphics[width=\linewidth]{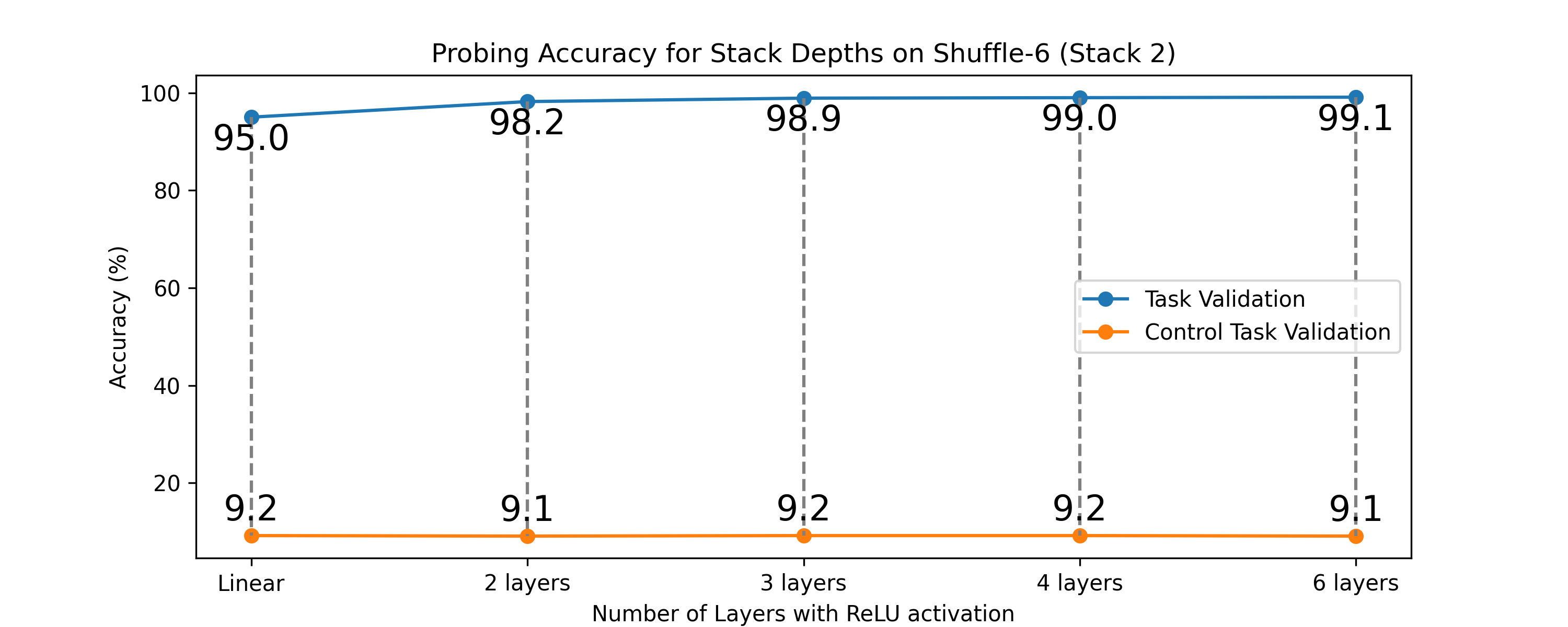}
    \caption{Probing accuracy across model architectures for different stack depths. Blue lines show task validation accuracy, while red lines represent a randomized control baseline. High task accuracy with low control accuracy indicates successful learning of stack-like structures.}
    \label{fig:vertical-stack}
\end{figure}

We probe the model trained on Dyck-1 for the stack depth at each input token, and for the models trained on Shuffle-$k$ we use $k$ probes - one for each stack.

We present the results for probing accuracies on Dyck-1, and one stack of the Shuffle languages in Figure \ref{fig:vertical-stack}, while the probing results for the rest of the stacks can be found in the Appendix. 

The probing accuracy, even in the case of linear probes, is high and since the accuracies on the control task are near random baselines it suggests that the probing models are not learning spurious connections or memorizing the training set. 

The results strongly hint towards the presence of counter variables in the internal representations of the models trained to recognize counter languages. 

Interestingly, the probe accuracy is significantly higher in Shuffle languages than Dyck-1, and within the Shuffle languages is higher for higher values of $k$. This can be attributed to the fact that as the number of stacks required increase the frequency at which their depths change decreases (for instance, the stack for Dyck-1 will get updated every token, but each stack in Shuffle-6 will get updated, on average, at every 6th token).
\section{Future Works and Conclusion}

A key limitation of our study is that probing classifiers do not provide any causal information about the feature they are used to detect. While the current work shows that the models learn counter-like structures, we leave to future work to determine the extent to which these structures provide a full picture of the data structures used by the model to compute the tasks, and to determine whether these structures play a causal role in determining the model's output.

The broader task of interpretability can be broken down into two sub-tasks - firstly, what representations and data structures a model learns, and secondly, what algorithms operate over, update, and read out these representations. This work only focuses on the former task, however the understanding of how these stack structures are updated is an important area of research.

It is also possible for the LM to learn multiple representations and algorithms, in which case exploring the causality of these becomes important. Moreover, ablating across multiple architecture choices like positional embedding type, downstream task (classifier vs next token prediction vs regression) etc. might be useful.

It is also useful to analyse the failure cases of the probing model and to see if there are specific kinds of strings on which the probing models misclassify the stack depths.

Since, our results show extremely positive signals for the presence of stacks, future work should also include the use of more robust techniques from MI techniques to confirm and extend these findings.

\bibliography{references}

\begin{thebibliography}{30}
\providecommand{\natexlab}[1]{#1}

\bibitem[{Abdou et~al.(2021)Abdou, Kulmizev, Hershcovich, Frank, Pavlick, and S{\o}gaard}]{abdou-etal-2021-language}
Mostafa Abdou, Artur Kulmizev, Daniel Hershcovich, Stella Frank, Ellie Pavlick, and Anders S{\o}gaard. 2021.
\newblock \href {https://doi.org/10.18653/v1/2021.conll-1.9} {Can language models encode perceptual structure without grounding? a case study in color}.
\newblock In \emph{Proceedings of the 25th Conference on Computational Natural Language Learning}, pages 109--132, Online. Association for Computational Linguistics.

\bibitem[{Ackerman and Cybenko(2020)}]{ackerman2020surveyneuralnetworksformal}
Joshua Ackerman and George Cybenko. 2020.
\newblock \href {https://arxiv.org/abs/2006.01338} {A survey of neural networks and formal languages}.
\newblock \emph{Preprint}, arXiv:2006.01338.

\bibitem[{Belinkov(2022)}]{belinkov-2022-probing}
Yonatan Belinkov. 2022.
\newblock \href {https://doi.org/10.1162/coli_a_00422} {Probing classifiers: Promises, shortcomings, and advances}.
\newblock \emph{Computational Linguistics}, 48(1):207--219.

\bibitem[{Bhattamishra et~al.(2020)Bhattamishra, Ahuja, and Goyal}]{Bhattamishra2020OnTA}
S.~Bhattamishra, Kabir Ahuja, and Navin Goyal. 2020.
\newblock \href {https://api.semanticscholar.org/CorpusID:222225236} {On the ability and limitations of transformers to recognize formal languages}.
\newblock In \emph{Conference on Empirical Methods in Natural Language Processing}.

\bibitem[{Coenen et~al.(2019)Coenen, Reif, Yuan, Kim, Pearce, Viégas, and Wattenberg}]{coenen2019visualizingmeasuringgeometrybert}
Andy Coenen, Emily Reif, Ann Yuan, Been Kim, Adam Pearce, Fernanda Viégas, and Martin Wattenberg. 2019.
\newblock \href {https://arxiv.org/abs/1906.02715} {Visualizing and measuring the geometry of bert}.
\newblock \emph{Preprint}, arXiv:1906.02715.

\bibitem[{DuSell and Chiang(2024)}]{stackattention}
Brian DuSell and David Chiang. 2024.
\newblock \href {https://arxiv.org/abs/2310.01749} {Stack attention: Improving the ability of transformers to model hierarchical patterns}.
\newblock \emph{Preprint}, arXiv:2310.01749.

\bibitem[{Elhage et~al.(2022)Elhage, Hume, Olsson, Schiefer, Henighan, Kravec, Hatfield-Dodds, Lasenby, Drain, Chen, Grosse, McCandlish, Kaplan, Amodei, Wattenberg, and Olah}]{elhage2022toymodelssuperposition}
Nelson Elhage, Tristan Hume, Catherine Olsson, Nicholas Schiefer, Tom Henighan, Shauna Kravec, Zac Hatfield-Dodds, Robert Lasenby, Dawn Drain, Carol Chen, Roger Grosse, Sam McCandlish, Jared Kaplan, Dario Amodei, Martin Wattenberg, and Christopher Olah. 2022.
\newblock \href {https://arxiv.org/abs/2209.10652} {Toy models of superposition}.
\newblock \emph{Preprint}, arXiv:2209.10652.

\bibitem[{Fernandez~Astudillo et~al.(2020)Fernandez~Astudillo, Ballesteros, Naseem, Blodgett, and Florian}]{transformerstack}
Ram{\'o}n Fernandez~Astudillo, Miguel Ballesteros, Tahira Naseem, Austin Blodgett, and Radu Florian. 2020.
\newblock \href {https://doi.org/10.18653/v1/2020.findings-emnlp.89} {Transition-based parsing with stack-transformers}.
\newblock In \emph{Findings of the Association for Computational Linguistics: EMNLP 2020}, pages 1001--1007, Online. Association for Computational Linguistics.

\bibitem[{Hahn(2020)}]{hahn-2020-theoretical}
Michael Hahn. 2020.
\newblock \href {https://doi.org/10.1162/tacl_a_00306} {Theoretical limitations of self-attention in neural sequence models}.
\newblock \emph{Transactions of the Association for Computational Linguistics}, 8:156--171.

\bibitem[{Hahn and Rofin(2024)}]{hahn-rofin-2024-sensitive}
Michael Hahn and Mark Rofin. 2024.
\newblock \href {https://doi.org/10.18653/v1/2024.acl-long.800} {Why are sensitive functions hard for transformers?}
\newblock In \emph{Proceedings of the 62nd Annual Meeting of the Association for Computational Linguistics (Volume 1: Long Papers)}, pages 14973--15008, Bangkok, Thailand. Association for Computational Linguistics.

\bibitem[{Hewitt and Liang(2019)}]{hewitt2019designinginterpretingprobescontrol}
John Hewitt and Percy Liang. 2019.
\newblock \href {https://arxiv.org/abs/1909.03368} {Designing and interpreting probes with control tasks}.
\newblock \emph{Preprint}, arXiv:1909.03368.

\bibitem[{J{\"a}ger and Rogers(2012)}]{Jger2012FormalLT}
Gerhard J{\"a}ger and James Rogers. 2012.
\newblock \href {https://api.semanticscholar.org/CorpusID:8765815} {Formal language theory: refining the chomsky hierarchy}.
\newblock \emph{Philosophical Transactions of the Royal Society B: Biological Sciences}, 367:1956 -- 1970.

\bibitem[{Joulin and Mikolov(2015)}]{stackrnn}
Armand Joulin and Tomas Mikolov. 2015.
\newblock \href {https://arxiv.org/abs/1503.01007} {Inferring algorithmic patterns with stack-augmented recurrent nets}.
\newblock \emph{Preprint}, arXiv:1503.01007.

\bibitem[{Kornai(1985)}]{kornai-1985-natural}
Andr{\'a}s Kornai. 1985.
\newblock \href {https://aclanthology.org/E85-1001/} {Natural languages and the {C}homsky hierarchy}.
\newblock In \emph{Second Conference of the {E}uropean Chapter of the Association for Computational Linguistics}, Geneva, Switzerland. Association for Computational Linguistics.

\bibitem[{Li et~al.(2021)Li, Nye, and Andreas}]{li-etal-2021-implicit}
Belinda~Z. Li, Maxwell Nye, and Jacob Andreas. 2021.
\newblock \href {https://doi.org/10.18653/v1/2021.acl-long.143} {Implicit representations of meaning in neural language models}.
\newblock In \emph{Proceedings of the 59th Annual Meeting of the Association for Computational Linguistics and the 11th International Joint Conference on Natural Language Processing (Volume 1: Long Papers)}, pages 1813--1827, Online. Association for Computational Linguistics.

\bibitem[{Li et~al.(2024)Li, Hopkins, Bau, Viégas, Pfister, and Wattenberg}]{li2024emergentworldrepresentationsexploring}
Kenneth Li, Aspen~K. Hopkins, David Bau, Fernanda Viégas, Hanspeter Pfister, and Martin Wattenberg. 2024.
\newblock \href {https://arxiv.org/abs/2210.13382} {Emergent world representations: Exploring a sequence model trained on a synthetic task}.
\newblock \emph{Preprint}, arXiv:2210.13382.

\bibitem[{Lindner et~al.(2023)Lindner, Kramar, Farquhar, Rahtz, McGrath, and Mikulik}]{NEURIPS2023_771155ab}
David Lindner, Janos Kramar, Sebastian Farquhar, Matthew Rahtz, Tom McGrath, and Vladimir Mikulik. 2023.
\newblock \href {https://proceedings.neurips.cc/paper_files/paper/2023/file/771155abaae744e08576f1f3b4b7ac0d-Paper-Conference.pdf} {Tracr: Compiled transformers as a laboratory for interpretability}.
\newblock In \emph{Advances in Neural Information Processing Systems}, volume~36, pages 37876--37899. Curran Associates, Inc.

\bibitem[{Nanda et~al.()Nanda, Chan, Lieberum, Smith, and Steinhardt}]{nandaprogress}
Neel Nanda, Lawrence Chan, Tom Lieberum, Jess Smith, and Jacob Steinhardt.
\newblock Progress measures for grokking via mechanistic interpretability.
\newblock In \emph{The Eleventh International Conference on Learning Representations}.

\bibitem[{Pérez et~al.(2019)Pérez, Marinković, and Barceló}]{pérez2019turingcompletenessmodernneural}
Jorge Pérez, Javier Marinković, and Pablo Barceló. 2019.
\newblock \href {https://arxiv.org/abs/1901.03429} {On the turing completeness of modern neural network architectures}.
\newblock \emph{Preprint}, arXiv:1901.03429.

\bibitem[{Rai et~al.(2024)Rai, Zhou, Feng, Saparov, and Yao}]{rai2024practicalreviewmechanisticinterpretability}
Daking Rai, Yilun Zhou, Shi Feng, Abulhair Saparov, and Ziyu Yao. 2024.
\newblock \href {https://arxiv.org/abs/2407.02646} {A practical review of mechanistic interpretability for transformer-based language models}.
\newblock \emph{Preprint}, arXiv:2407.02646.

\bibitem[{Rogers et~al.(2020)Rogers, Kovaleva, and Rumshisky}]{rogers2020primer}
Anna Rogers, Olga Kovaleva, and Anna Rumshisky. 2020.
\newblock \href {https://doi.org/10.1162/tacl_a_00349} {A primer in {BERT}ology: What we know about how {BERT} works}.
\newblock \emph{Transactions of the Association for Computational Linguistics}, 8:842--866.

\bibitem[{Strobl et~al.(2024{\natexlab{a}})Strobl, Merrill, Weiss, Chiang, and Angluin}]{strobl-etal-2024-formal}
Lena Strobl, William Merrill, Gail Weiss, David Chiang, and Dana Angluin. 2024{\natexlab{a}}.
\newblock \href {https://doi.org/10.1162/tacl_a_00663} {What formal languages can transformers express? a survey}.
\newblock \emph{Transactions of the Association for Computational Linguistics}, 12:543--561.

\bibitem[{Strobl et~al.(2024{\natexlab{b}})Strobl, Merrill, Weiss, Chiang, and Angluin}]{Strobl_2024}
Lena Strobl, William Merrill, Gail Weiss, David Chiang, and Dana Angluin. 2024{\natexlab{b}}.
\newblock \href {https://doi.org/10.1162/tacl_a_00663} {What formal languages can transformers express? a survey}.
\newblock \emph{Transactions of the Association for Computational Linguistics}, 12:543–561.

\bibitem[{Suzgun et~al.(2019)Suzgun, Gehrmann, Belinkov, and Shieber}]{suzgun2019memoryaugmentedrecurrentneuralnetworks}
Mirac Suzgun, Sebastian Gehrmann, Yonatan Belinkov, and Stuart~M. Shieber. 2019.
\newblock \href {https://arxiv.org/abs/1911.03329} {Memory-augmented recurrent neural networks can learn generalized dyck languages}.
\newblock \emph{Preprint}, arXiv:1911.03329.

\bibitem[{Vafa et~al.(2024)Vafa, Chen, Rambachan, Kleinberg, and Mullainathan}]{vafa2024evaluatingworldmodelimplicit}
Keyon Vafa, Justin~Y. Chen, Ashesh Rambachan, Jon Kleinberg, and Sendhil Mullainathan. 2024.
\newblock \href {https://arxiv.org/abs/2406.03689} {Evaluating the world model implicit in a generative model}.
\newblock \emph{Preprint}, arXiv:2406.03689.

\bibitem[{Voita et~al.(2019)Voita, Talbot, Moiseev, Sennrich, and Titov}]{voita2019analyzingattention}
Elena Voita, David Talbot, Fedor Moiseev, Rico Sennrich, and Ivan Titov. 2019.
\newblock \href {https://doi.org/10.18653/v1/P19-1580} {Analyzing multi-head self-attention: Specialized heads do the heavy lifting, the rest can be pruned}.
\newblock In \emph{Proceedings of the 57th Annual Meeting of the Association for Computational Linguistics}, pages 5797--5808, Florence, Italy. Association for Computational Linguistics.

\bibitem[{Weiss et~al.(2021)Weiss, Goldberg, and Yahav}]{weiss2021thinking}
Gail Weiss, Yoav Goldberg, and Eran Yahav. 2021.
\newblock Thinking like transformers.
\newblock In \emph{International Conference on Machine Learning}, pages 11080--11090. PMLR.

\bibitem[{Yun et~al.(2020)Yun, Bhojanapalli, Rawat, Reddi, and Kumar}]{yun2020transformersuniversalapproximatorssequencetosequence}
Chulhee Yun, Srinadh Bhojanapalli, Ankit~Singh Rawat, Sashank~J. Reddi, and Sanjiv Kumar. 2020.
\newblock \href {https://arxiv.org/abs/1912.10077} {Are transformers universal approximators of sequence-to-sequence functions?}
\newblock \emph{Preprint}, arXiv:1912.10077.

\bibitem[{Zhang et~al.(2024)Zhang, Tigges, Zhang, Biderman, Raginsky, and Ringer}]{zhang2024transformerbasedmodelsperfectlearning}
Dylan Zhang, Curt Tigges, Zory Zhang, Stella Biderman, Maxim Raginsky, and Talia Ringer. 2024.
\newblock \href {https://arxiv.org/abs/2401.12947} {Transformer-based models are not yet perfect at learning to emulate structural recursion}.
\newblock \emph{Preprint}, arXiv:2401.12947.

\bibitem[{Zhou et~al.(2023)Zhou, Bradley, Littwin, Razin, Saremi, Susskind, Bengio, and Nakkiran}]{zhou2023algorithmstransformerslearnstudy}
Hattie Zhou, Arwen Bradley, Etai Littwin, Noam Razin, Omid Saremi, Josh Susskind, Samy Bengio, and Preetum Nakkiran. 2023.
\newblock \href {https://arxiv.org/abs/2310.16028} {What algorithms can transformers learn? a study in length generalization}.
\newblock \emph{Preprint}, arXiv:2310.16028.

\end{thebibliography}

\clearpage 
\appendix
\renewcommand{\thesection}{\Alph{section}}
\section{Counter Languages}
\label{sec:appendix_example}

In this section we formalize the notion of counter languages by defining them as languages modeled by a $k$-counter machine. For \( m \in \mathbb{Z} \), let \( \pm m \) denote the function \( \lambda x.\, x \pm m \). Let \( \times 0 \) denote the constant zero function \( \lambda x.\, 0 \). The $k$-counter machine is defined as \footnote{Merrill, W. (2020). On the linguistic capacity of real-time counter automata. arXiv preprint arXiv:2004.06866.}:
\\
\textbf{Definition 1 (General counter machine).} A $k$-counter machine is a tuple $\langle \Sigma, Q, q_0, u, \delta, F \rangle$ with
\begin{enumerate}
    \item A finite alphabet $\Sigma$
    \item A finite set of states $Q$
    \item An initial state $q_0$
    \item A counter update function
    \begin{align*}
        u : \Sigma \times Q \times& \{0,1\}^k \to \\
        &(\{+m : m \in \mathbb{Z}\} \cup \{\times 0\})^k
    \end{align*}
    \item A state transition function
    \[
        \delta : \Sigma \times Q \times \{0,1\}^k \to Q
    \]
    \item An acceptance mask
    \[
        F \subseteq Q \times \{0,1\}^k
    \]
\end{enumerate}

A machine processes an input string $x$ one token at a time. For each token, we use $u$ to update the counters and $\delta$ to update the state according to the current input token, the current state, and a finite mask of the current counter values. We formalize this in Definition 2.

For a vector $\mathbf{v}$, let $z(\mathbf{v})$ denote the broadcasted ``zero-check" function, i.e.
\[
    z(\mathbf{v})_i =
    \begin{cases}
        0 & \text{if } v_i = 0 \\
        1 & \text{otherwise}.
    \end{cases}
\]

\textbf{Definition 2 (Counter machine computation).} Let $\langle q, c \rangle \in Q \times \mathbb{Z}^k$ be a configuration of machine $M$. Upon reading the input $x_t \in \Sigma$, we define the transition
\[
    \langle q, c \rangle \to_{x_t} \langle \delta(x_t, q, z(c)), u(x_t, q, z(c))(c) \rangle.
\]

\textbf{Definition 3 (Real-time acceptance).} For any string $x \in \Sigma^*$ with length $n$, a counter machine accepts $x$ if there exist states $q_1, \dots, q_n$ and counter configurations $c_1, \dots, c_n$ such that
\[
    \langle q_0, 0 \rangle \to_{x_1} \langle q_1, c_1 \rangle \to_{x_2} \dots \to_{x_n} \langle q_n, c_n \rangle \in F.
\]

\textbf{Definition 4 (Real-time language acceptance).} A counter machine accepts a language $L$ if, for each $x \in \Sigma^*$, it accepts $x$ iff $x \in L$.

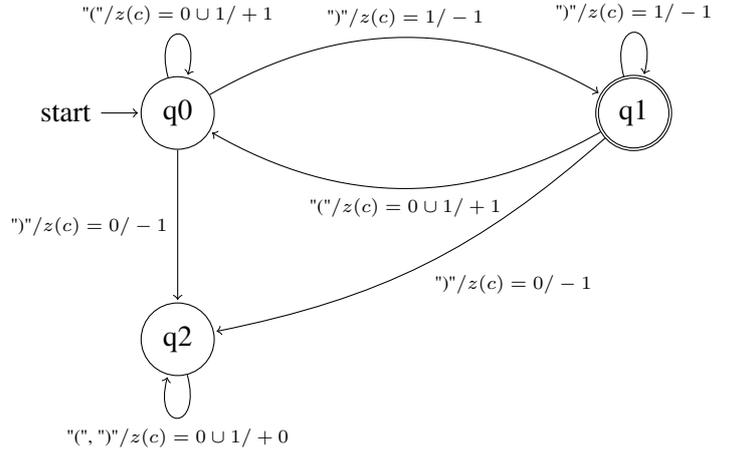
\begin{figure}[H]
    \centering
    \begin{tikzpicture}[shorten >=1pt, node distance=3cm, on grid, auto] 
        \node[state, initial] (q0) {q0}; 
        \node[state, accepting] (q1) [right=6cm of q0] {q1}; 
        \node[state] (q2) [below=3cm of q0] {q2};  

        \path[->] 
            (q0) edge [loop above] node{\scriptsize $\text{"("}/z(c)=0\cup1/+1$} (q0)
                 edge [bend left] node {\scriptsize $\text{")"}/z(c)=1/-1$} (q1)
                 edge [below] node[anchor = east] {\scriptsize $\text{")"}/z(c)=0/-1$} (q2)
            (q1) edge [loop above] node{\scriptsize $\text{")"}/z(c)=1/-1$} (q1)
                 edge [bend left] node {\scriptsize $\text{"("}/z(c)=0\cup1/+1$} (q0)
                 edge [bend left=15] node {\scriptsize $\text{")"}/z(c)=0/-1$} (q2)
            (q2) edge [loop below] node {\scriptsize $\text{"(", ")"}/z(c)=0\cup1/+0$} (q2);
    \end{tikzpicture}
    \caption{A graphical representation of a 1-counter machine that accepts Dyck-1 if we set $F$ to verify that the counter is 0 and we are in q1.}
    \label{fig:automaton}
\end{figure}

\vspace{-2mm} 

\begin{figure}[H]
    \centering
    $$
\langle 0, q_0 \rangle \xrightarrow[(]{} \langle 1, q_0 \rangle \xrightarrow[(]{} \langle 2, q_0 \rangle \xrightarrow[)]{} \langle 1, q_1 \rangle \xrightarrow[)]{} \langle 0, q_1 \rangle \in F 
$$
\\
$$
\langle 0, q_0 \rangle \xrightarrow[(]{} \langle 1, q_0 \rangle \xrightarrow[(]{} \langle 2, q_0 \rangle \xrightarrow[)]{} \langle 1, q_1 \rangle \xrightarrow[(]{} \langle 2, q_0 \rangle \notin F
$$
    \caption{Behavior of the counter machine in \ref{fig:automaton} on $(())$ (\textbf{top}) and $(()($ (\textbf{bottom})}

    \label{fig:counter-behavior}
\end{figure}



\section{Tracr Model Results}

As mentioned in \ref{sec:probing} we compile a model using Tracr from the following RASP program for Dyck-1\footnote{Weiss, Gail, Yoav Goldberg, and Eran Yahav. "Thinking like transformers." International Conference on Machine Learning. PMLR, 2021.}:
\begin{lstlisting}
def num_prevs(bools) {
    prevs = select(indices, indices, <=);
    return (indices+1) * 
           aggregate(prevs,
                     indicator(bools));
}

n_opens = num_prevs(tokens=="(");
n_closes = num_prevs(tokens==")");
balance = n_opens - n_closes;
prev_imbalances = num_prevs(balance<0);
dyck1 = "F" if prev_imbalances > 0 
           else 
           ("T" if balance==0 else "P");
\end{lstlisting}

Here, \textcolor{red}{P} refers to the state when the prefix at the current token is legal but not yet balances, \textcolor{red}{T} when it is balanced, and \textcolor{red}{F} when it is illegal. This process produces a transformer model which is then used to generate a probing dataset. We leverage the same training tokens used in previous experiments to generate activation embeddings, which are then utilized to train our probing model. To robustly demonstrate the effectiveness of our probing methodology, we train the model on two distinct tasks: multi-class classification and regression. Given the transformer's unique architectural design, which is tailored to align with the underlying algorithm, our probing approach aims to recover stack-like properties inherent in the language. If successful, this would provide conclusive evidence of the efficacy of our probing setup. By successfully recovering these properties—such as stack depth—through high accuracy on the probing tasks, we demonstrate the efficacy of our probing setup. These results provide strong evidence that our approach effectively captures the hierarchical and structural features encoded in the model's representations.

\begin{figure}[h] 
    \centering
    \includegraphics[width=\linewidth]{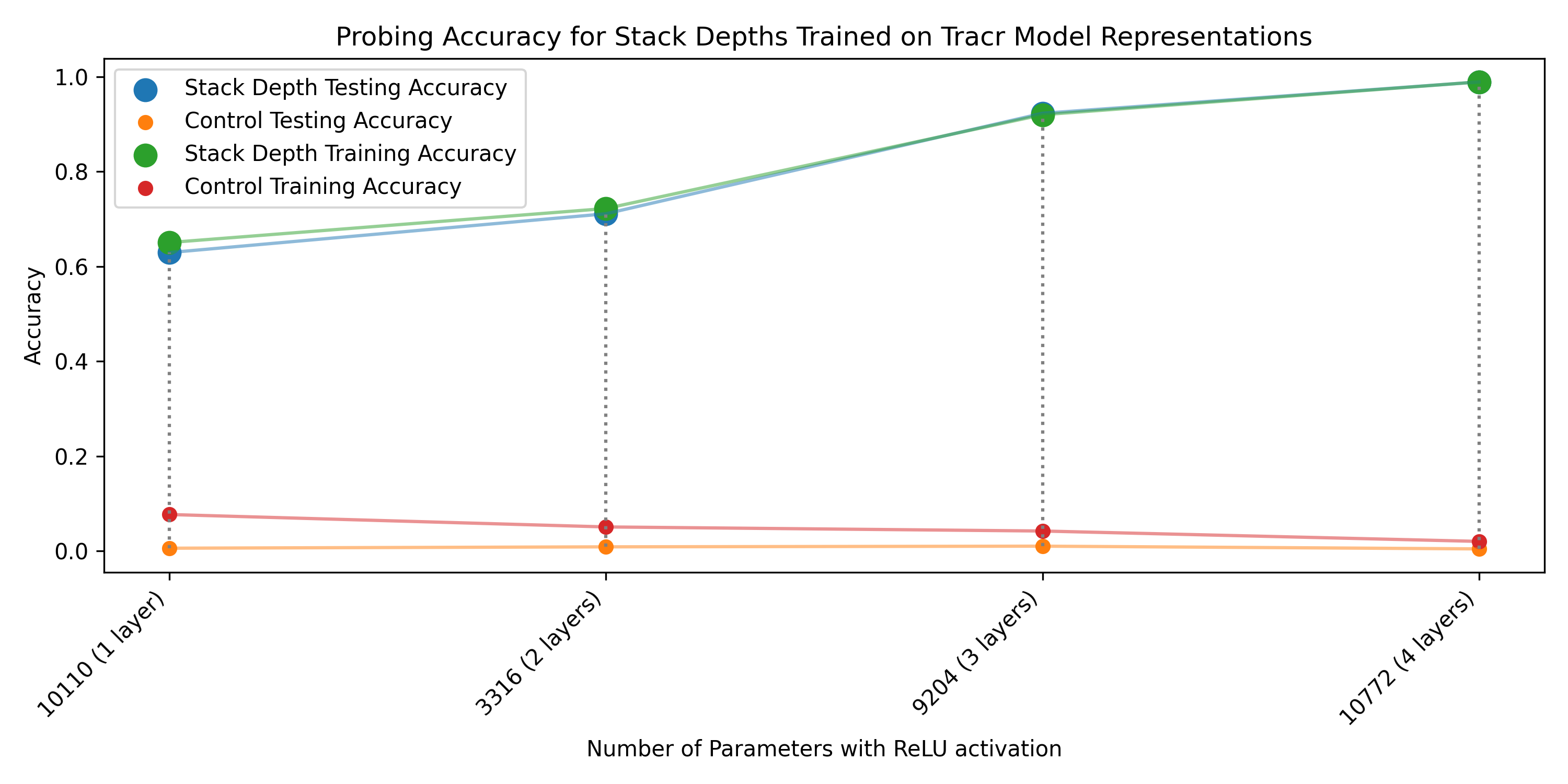} 
    \vskip\baselineskip 
    \includegraphics[width=\linewidth]{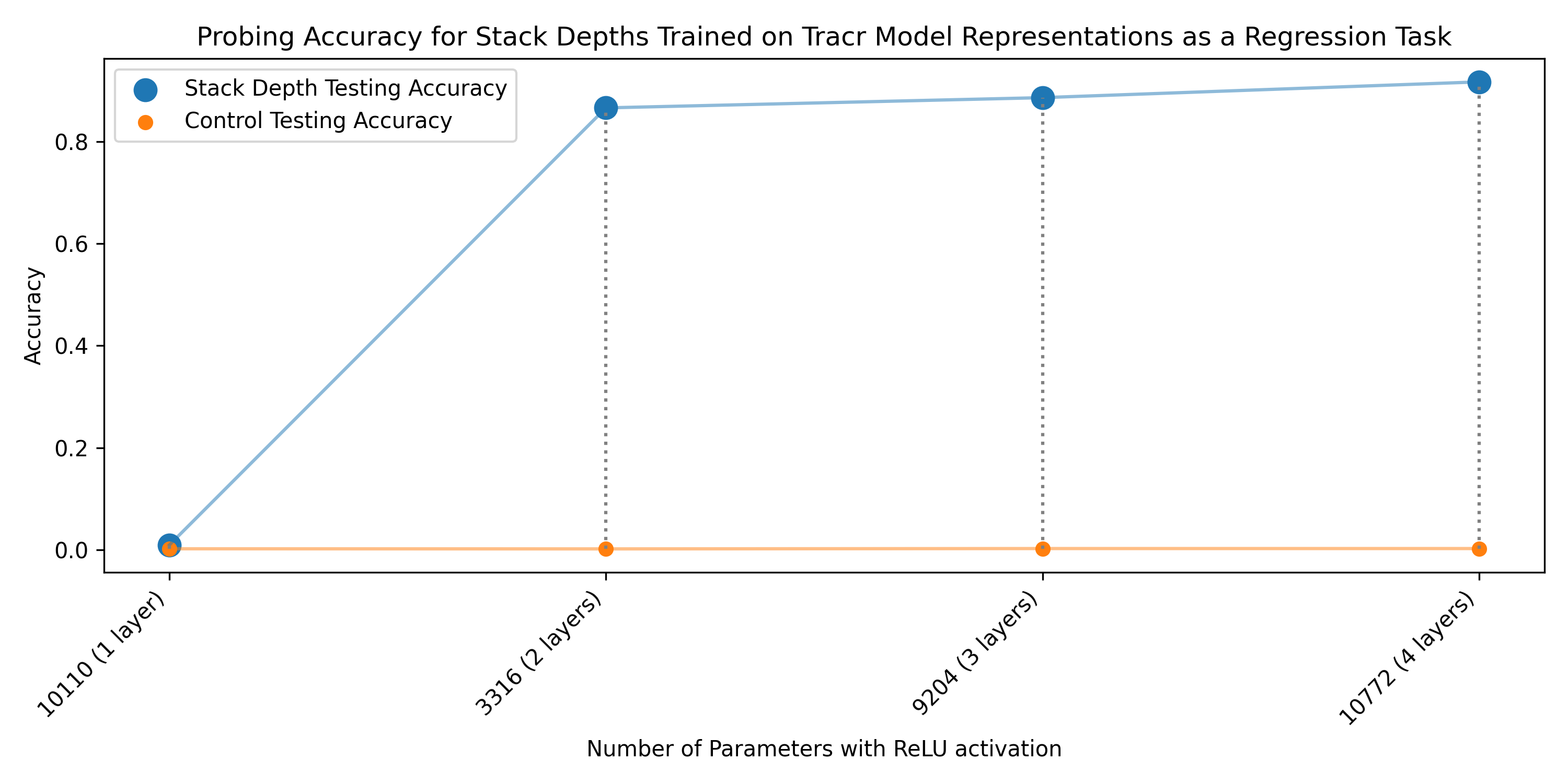}
    \caption{Probing results on the Tracr compiled model.}
    \label{fig:tracr}
\end{figure}


\vspace{3cm}
\section{Probing Model Results}

\subsection{Dyck-1}
\begin{figure}[h]
    \centering
    \includegraphics[width=\linewidth]{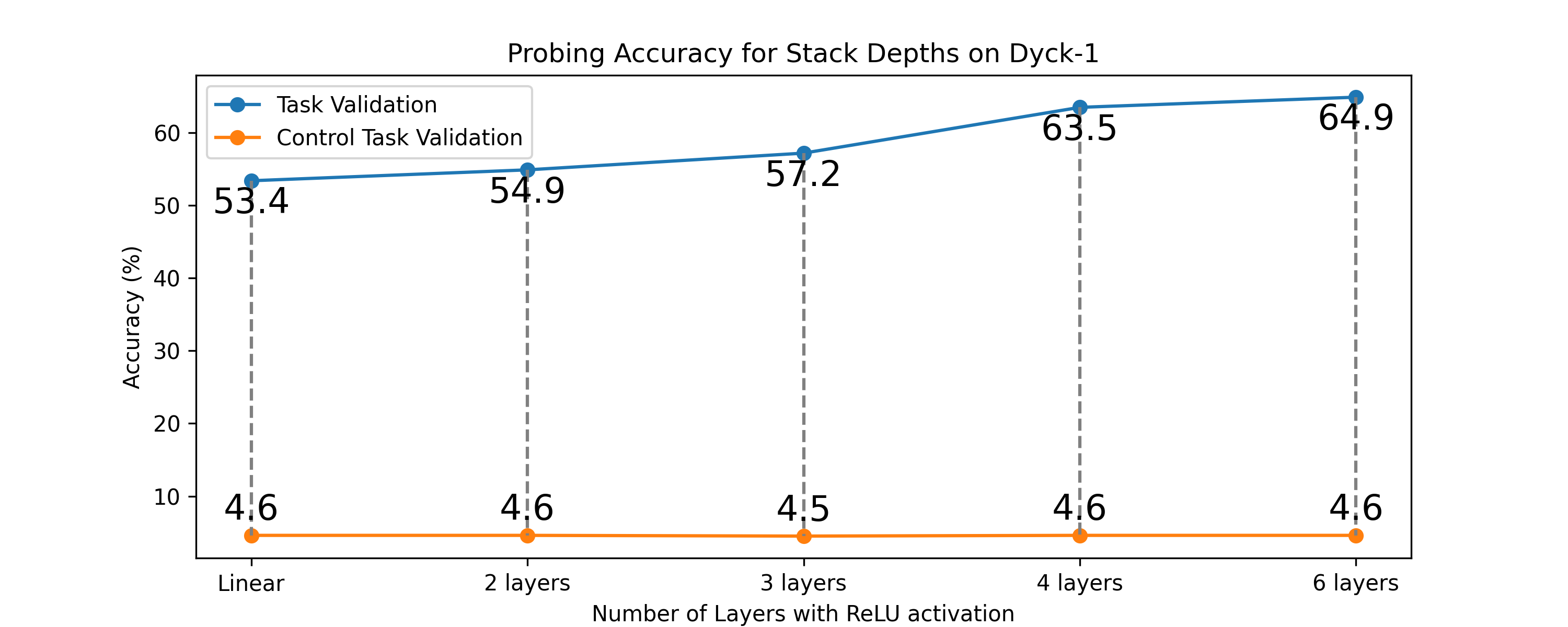}
    \label{fig:example}
\end{figure}

\subsection{Shuffle-2}
\begin{figure}[h]
    \centering
    \includegraphics[width=\linewidth]{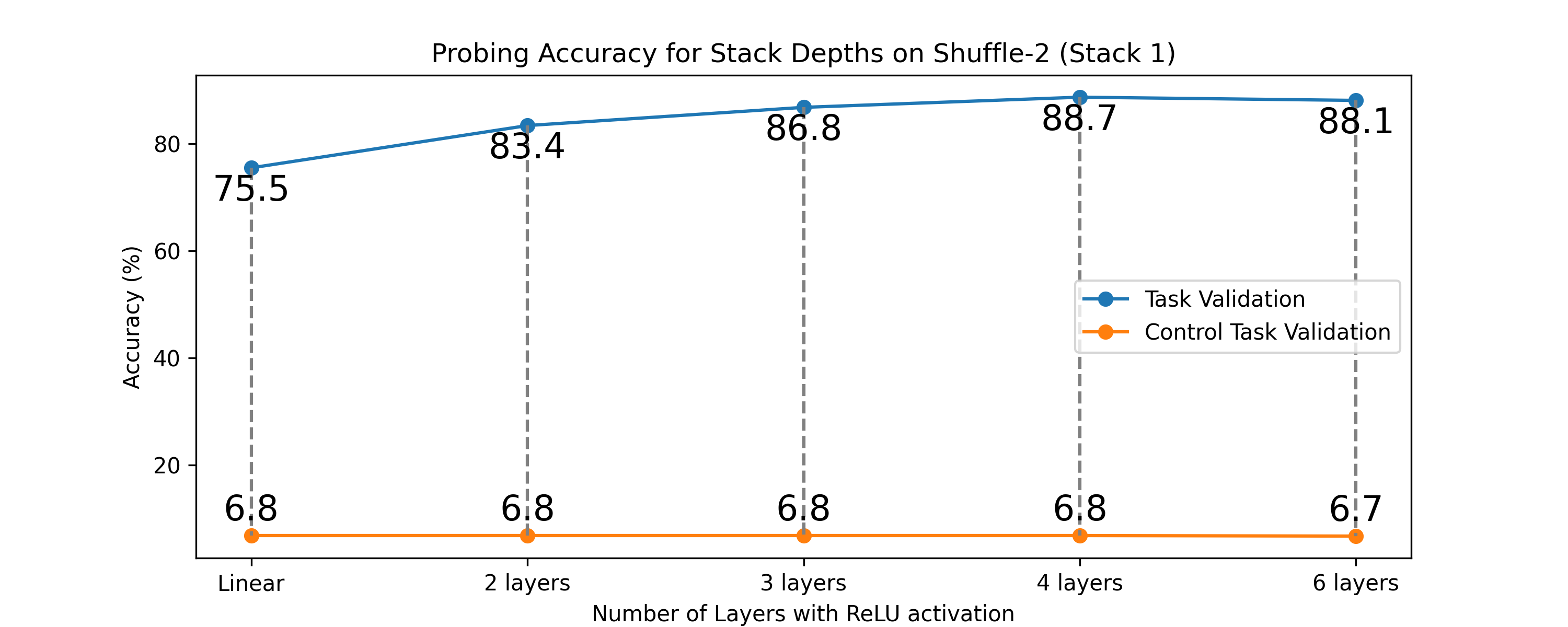}
    \label{fig:example}
\end{figure}
\begin{figure}[h]
    \centering
    \includegraphics[width=\linewidth]{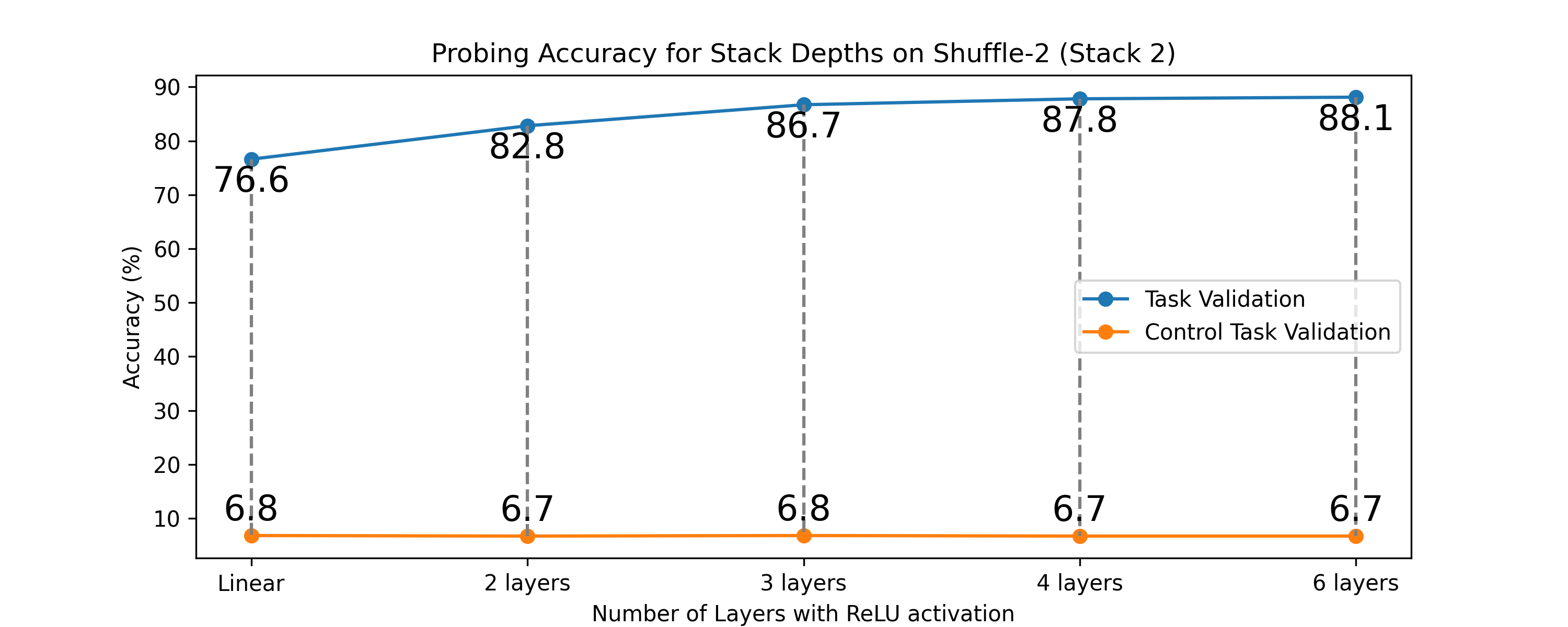}
    \label{fig:example}
\end{figure}

\subsection{Shuffle-4}

\begin{figure}[h!]
    \centering
    \includegraphics[width=\linewidth]{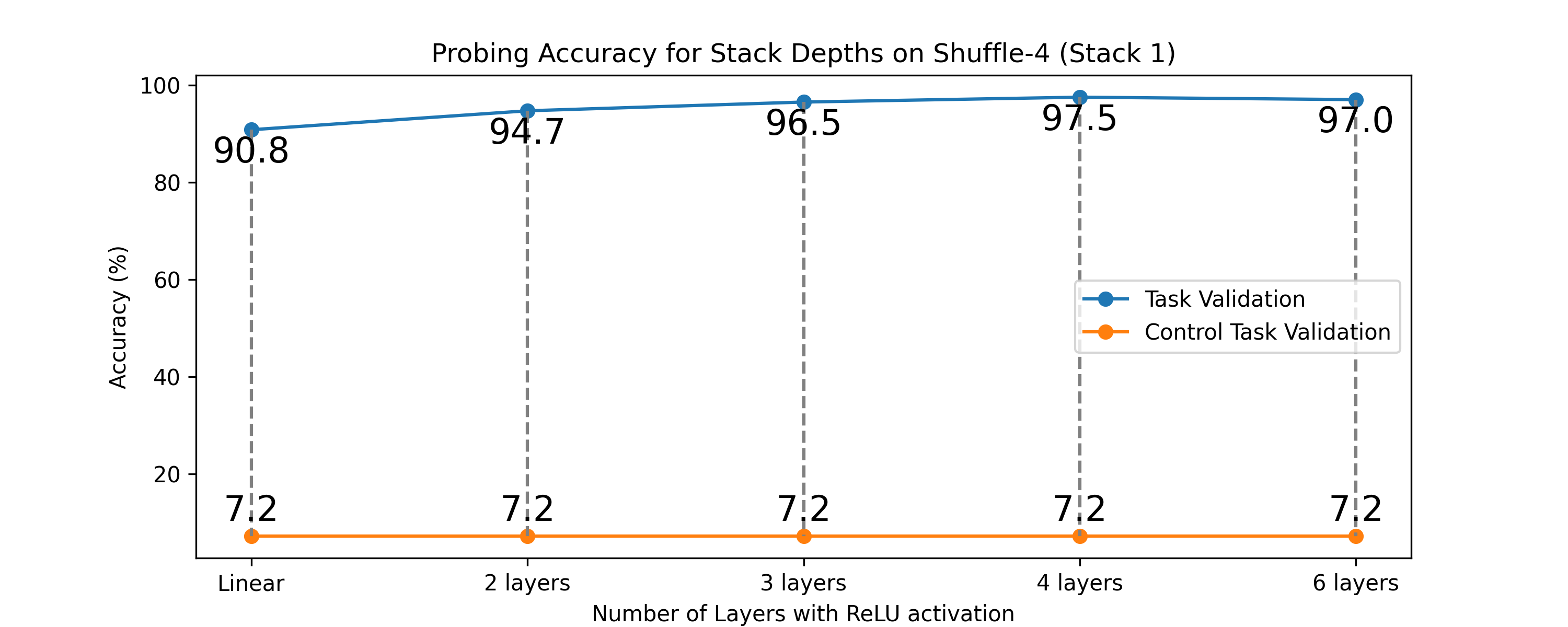}
    \label{fig:shuffle1}
\end{figure}

\begin{figure}[h!]
    \centering
    \includegraphics[width=\linewidth]{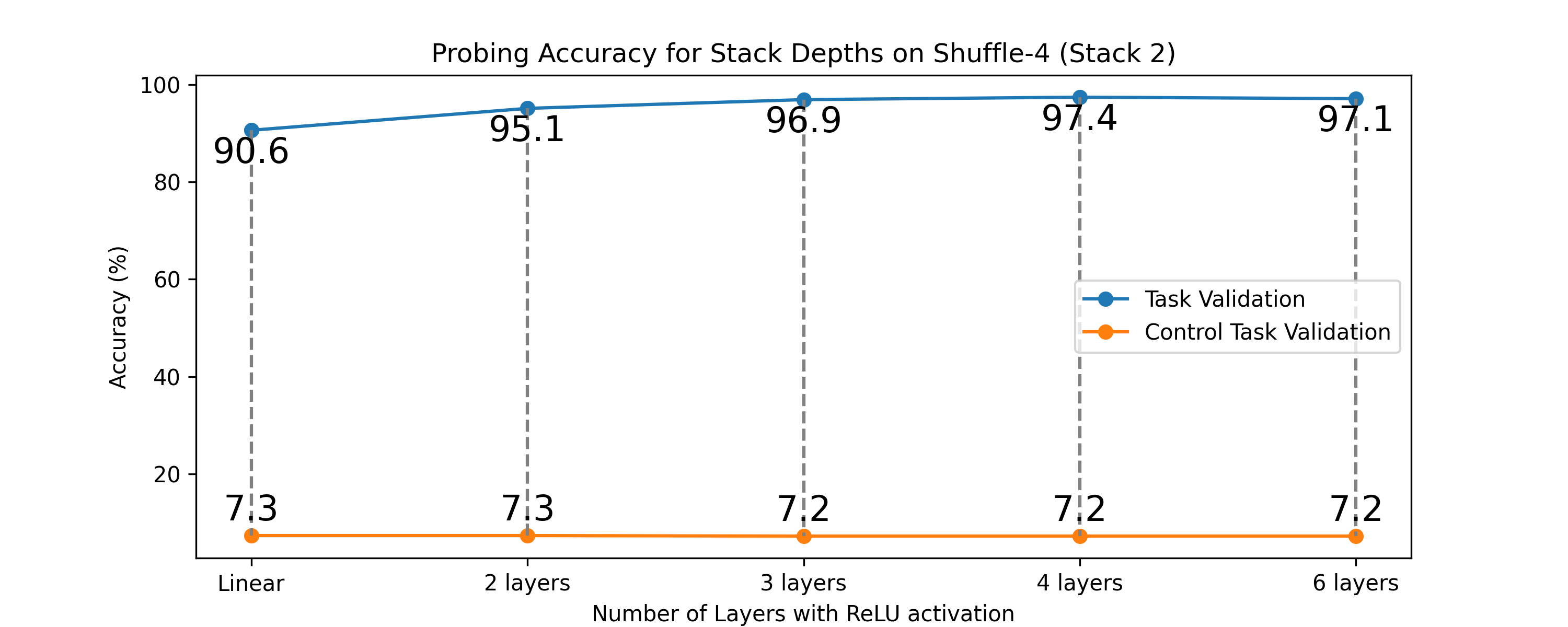}
    \label{fig:shuffle2}
\end{figure}

\begin{figure}[h!]
    \centering
    \includegraphics[width=\linewidth]{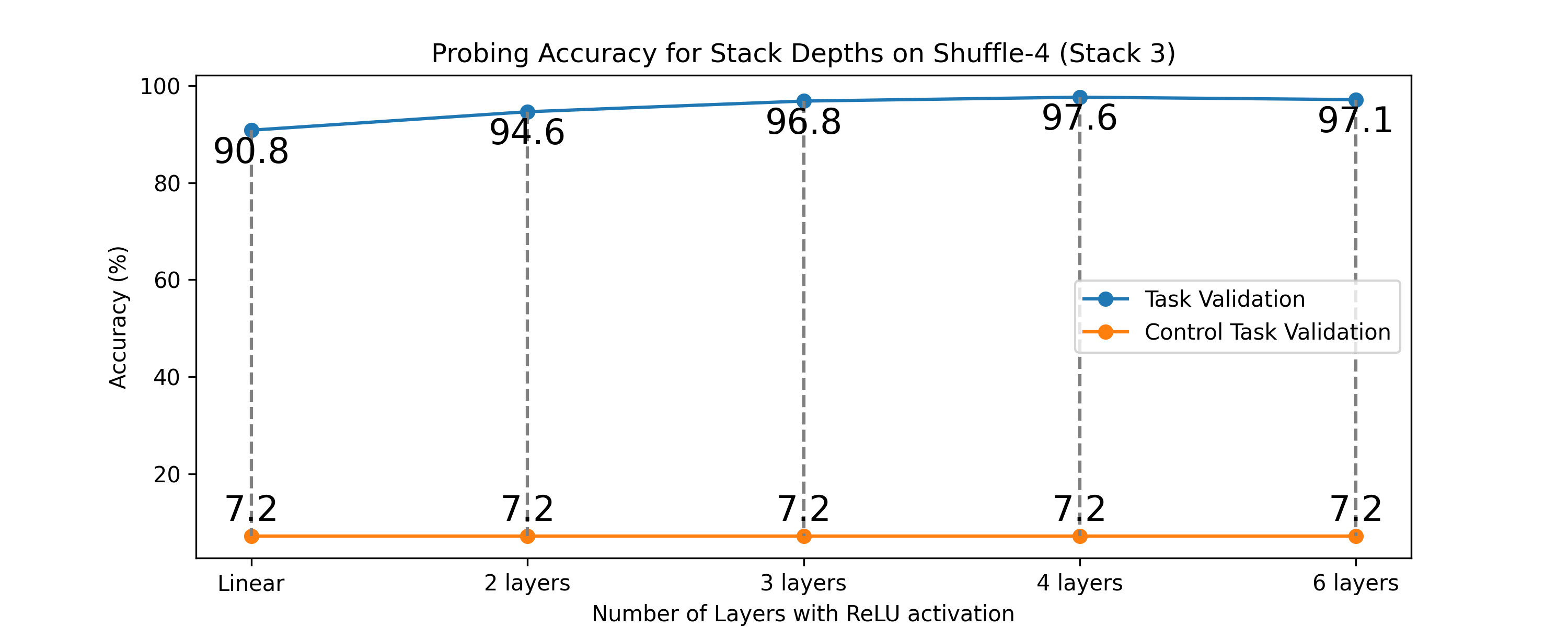}
    \label{fig:shuffle3}
\end{figure}

\begin{figure}[h!]
    \centering
    \includegraphics[width=\linewidth]{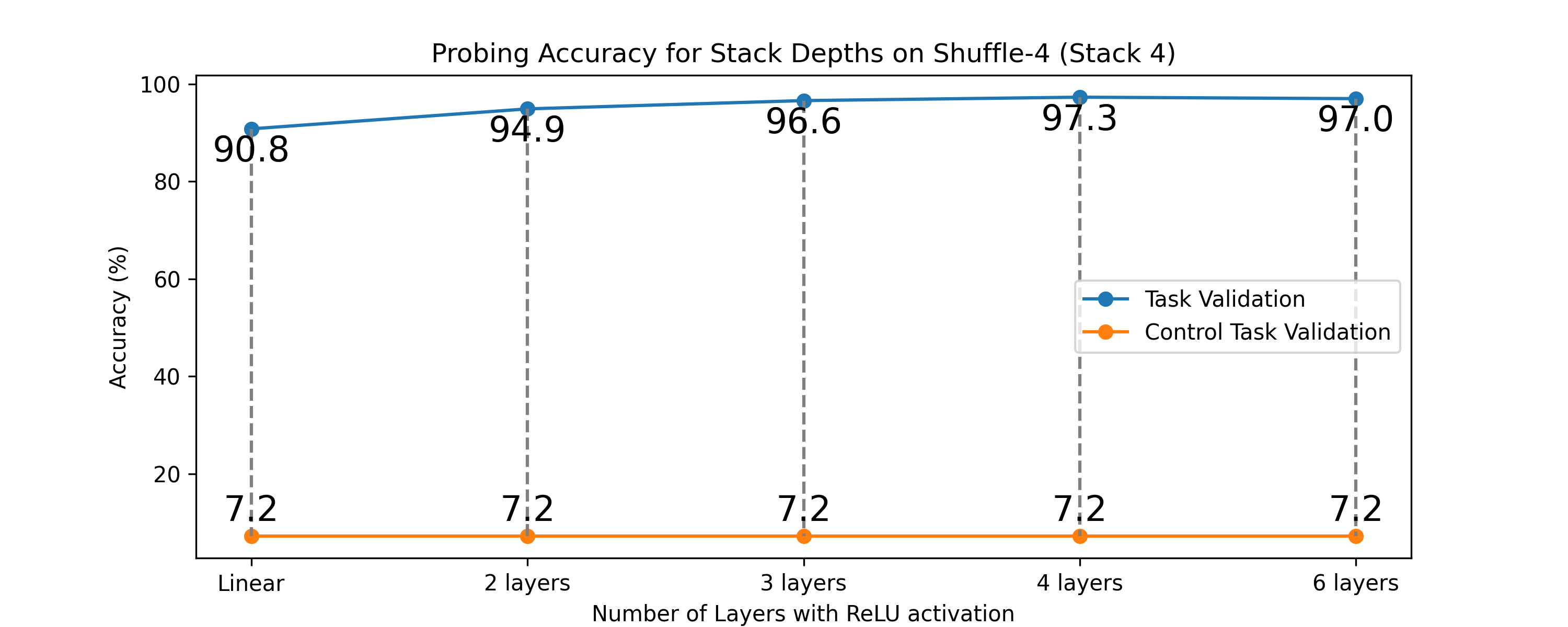}
    \label{fig:shuffle4}
\end{figure}

\newpage

\subsection{Shuffle-6}

\begin{figure}[h!]
    \centering
    \includegraphics[width=\linewidth]{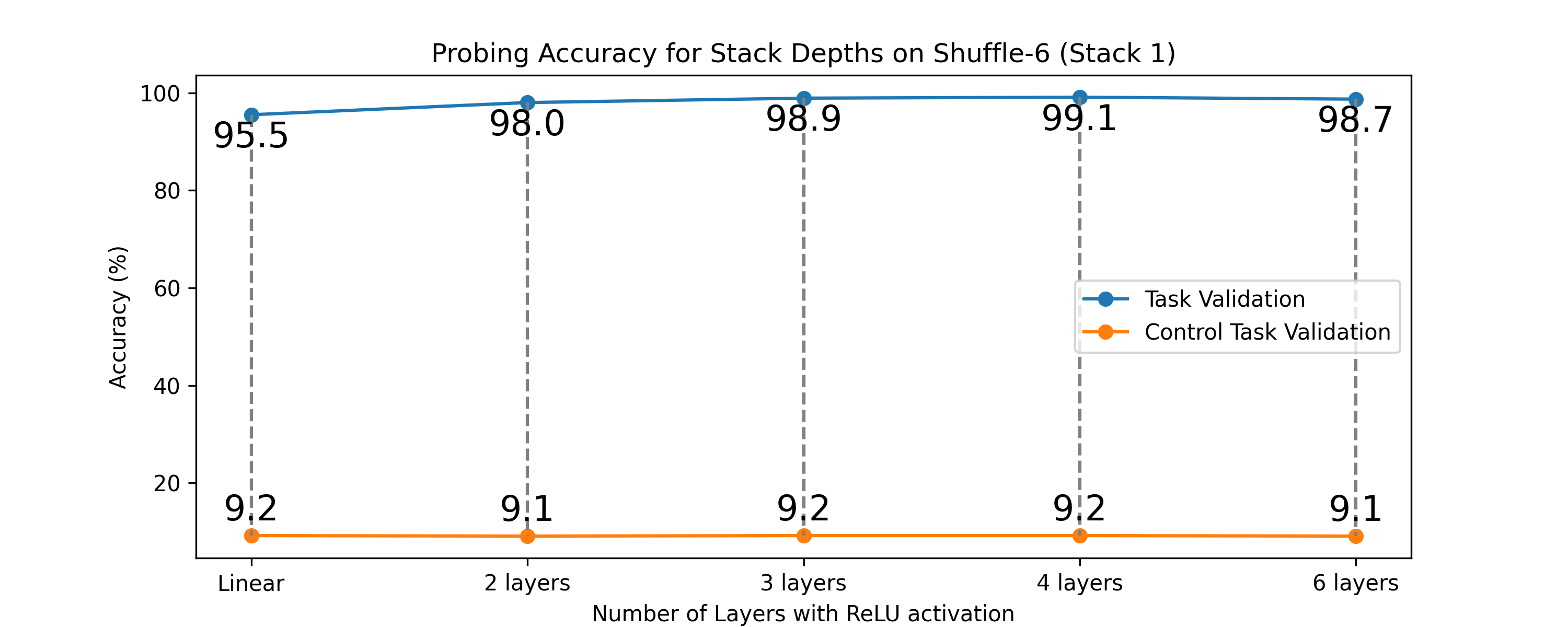}
    \label{fig:shuffle6_stack1}
\end{figure}

\begin{figure}[h!]
    \centering
    \includegraphics[width=\linewidth]{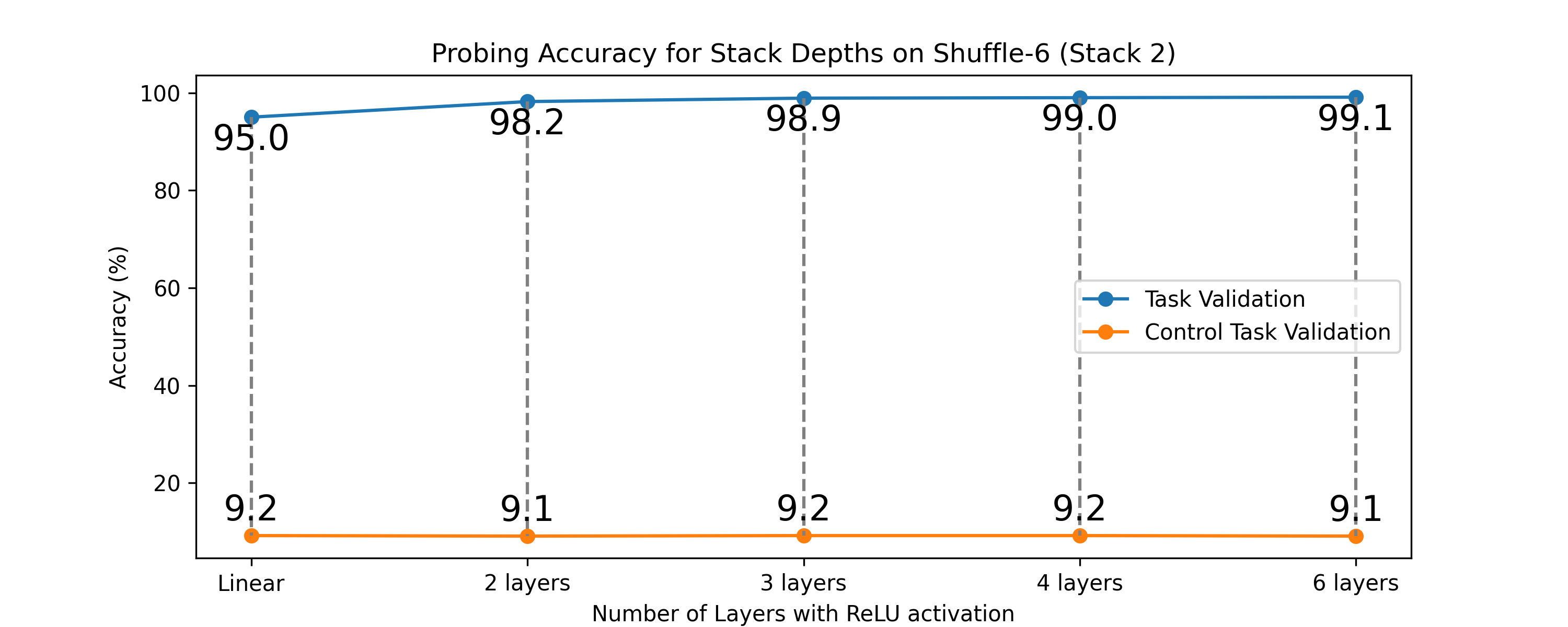}
    \label{fig:shuffle6_stack2}
\end{figure}

\begin{figure}[h!]
    \centering
    \includegraphics[width=\linewidth]{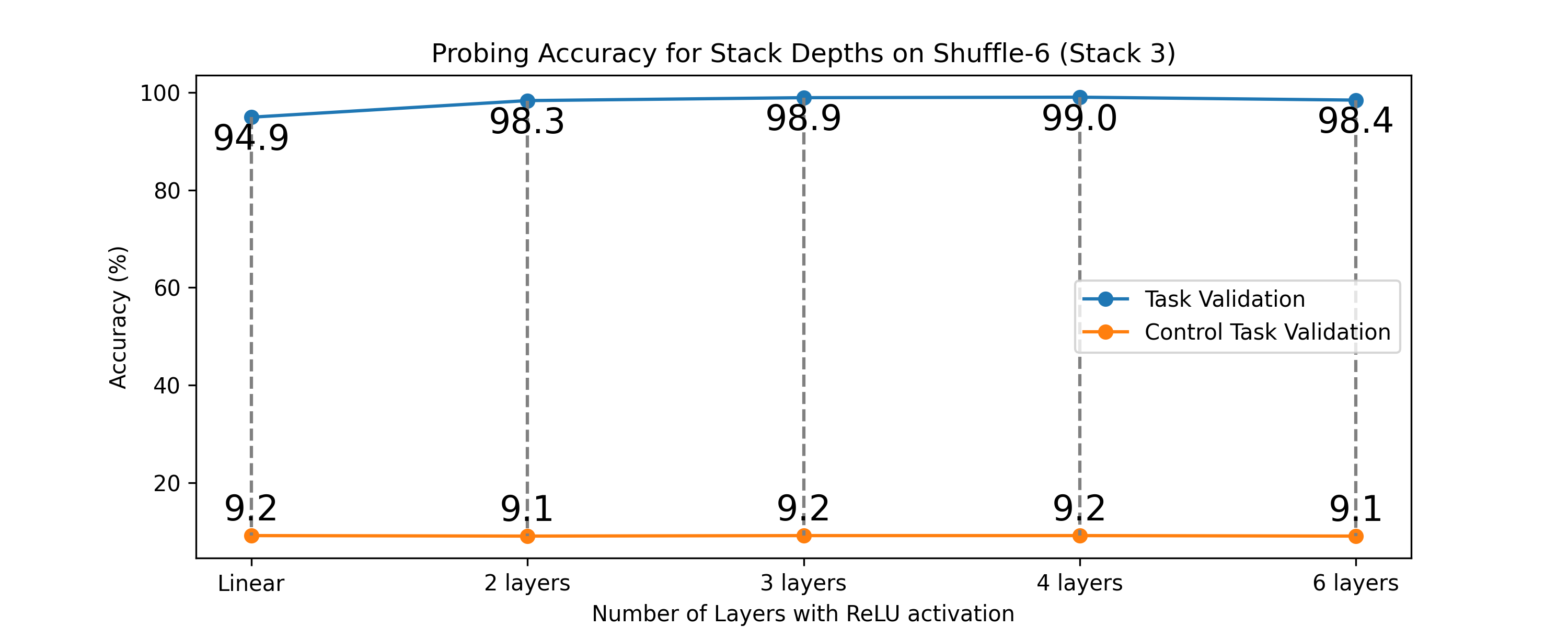}
    \label{fig:shuffle6_stack3}
\end{figure}

\begin{figure}[h!]
    \centering
    \includegraphics[width=\linewidth]{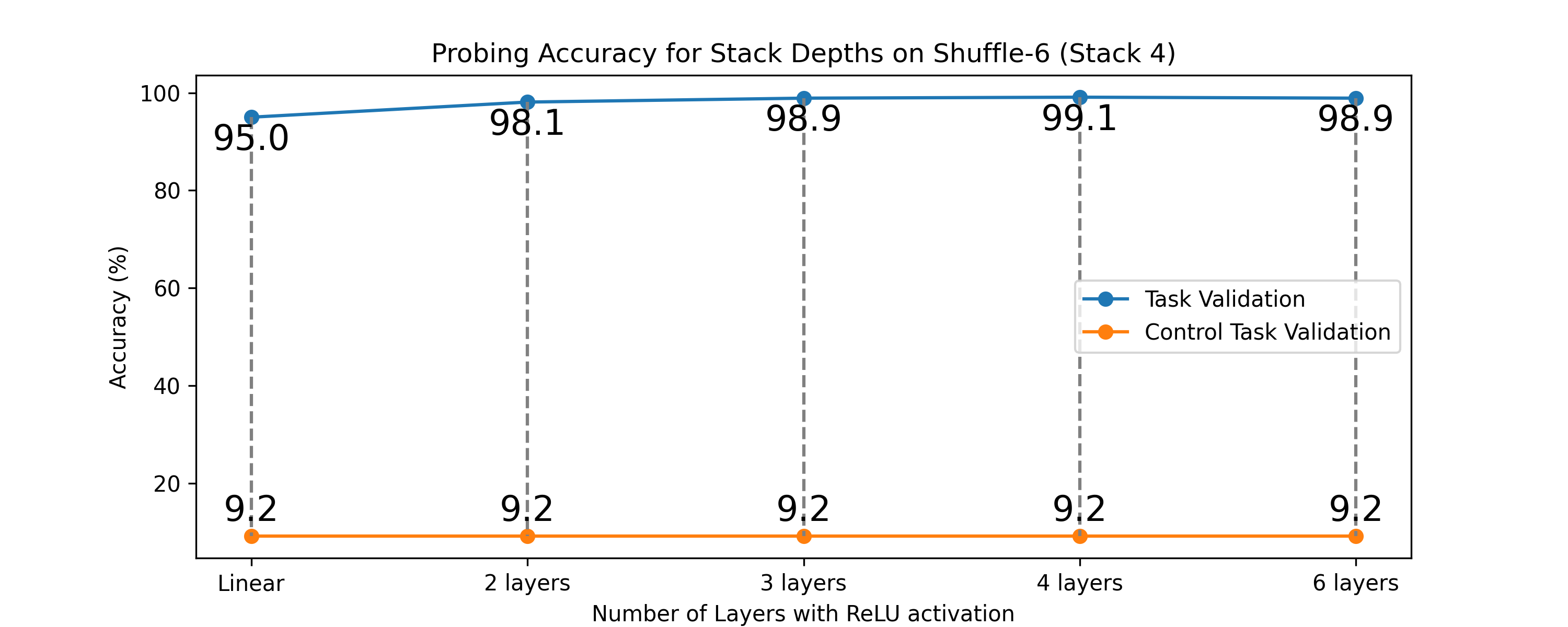}
    \label{fig:shuffle6_stack4}
\end{figure}

\begin{figure}[h!]
    \centering
    \includegraphics[width=\linewidth]{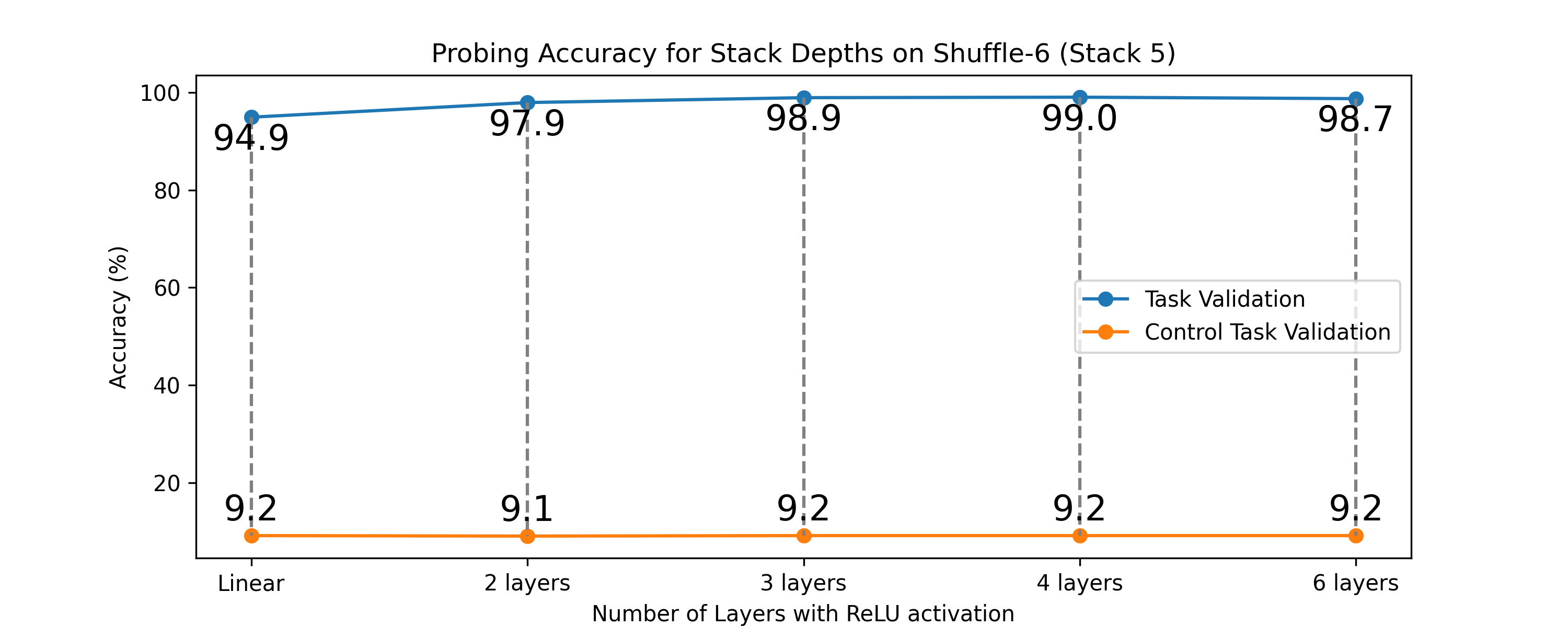}
    \label{fig:shuffle6_stack5}
\end{figure}

\begin{figure}[h!]
    \centering
    \includegraphics[width=\linewidth]{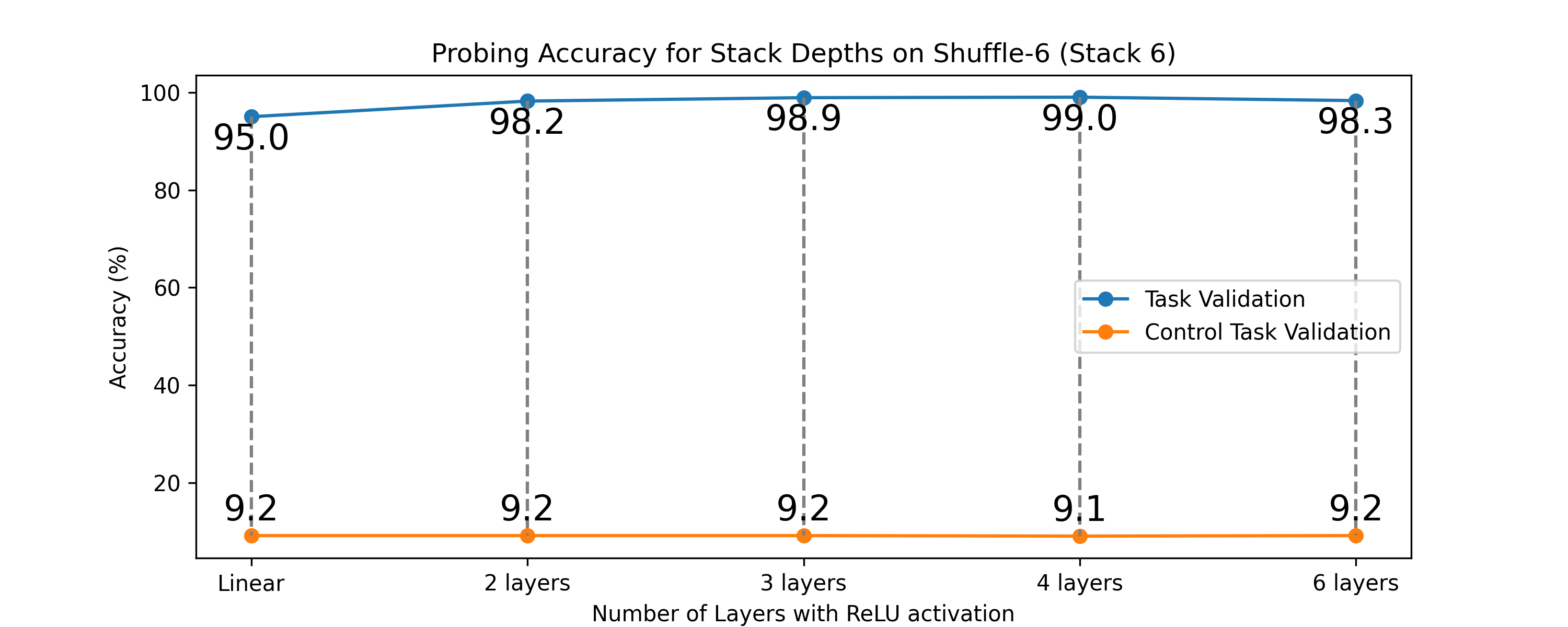}
    \label{fig:shuffle6_stack6}
\end{figure}

\section{Probing Model Training Details}

The probing classifier models are fully connected layers (ranging from linear models, to models with 6 layers with ReLU activations with hidden layer size of 128). We use dropout=$0.2$ and initialize the weights using Xavier initialization. The models are trained for 10 epochs with a learning rate of 0.001, batch size 32 with the Adam optimizer. We utilise the Cross-Entropy loss to train the multi-class classification task. 

\end{document}